# Quantifying Marketing Performance at Channel-Partner Level by Using Marketing Mix Modeling (MMM) and Shapley Value Regression

Sean Tang[*,†]  Sriya Musunuru[*,‡]  Baoshi Zong[*,§]  Brooks Thornton[*,¶]

Jan 2nd, 2024

**ABSTRACT**

This paper explores the application of Shapley Value Regression in dissecting marketing performance at channel-partner level, complementing channel-level Marketing Mix Modeling (MMM). Utilizing real-world data from the financial services industry, we demonstrate the practicality of Shapley Value Regression in evaluating individual partner contributions. Although structured in-field testing along with cooperative game theory is most accurate, it can often be highly complex and expensive to conduct. Shapley Value Regression is thus a more feasible approach to disentangle the influence of each marketing partner within a marketing channel. We also propose a simple method to derive adjusted coefficients of Shapley Value Regression and compare it with alternative approaches.

[*]Advanced Media Measurement Team, Discover Financial Services Inc., Riverwoods, IL
[†]xiaotang@discover.com

[‡]sriyamusunuru@discover.com

[§]baoshizong@discover.com

[¶]brooksthornton@discover.com

# 1. Introduction

In the ever-evolving marketing landscape, organizations of varying sizes leverage a mix of online and offline channels to get their marketing messages across. Though this maximizes the outreach of the audience, it is often a challenging task to accurately disentangle the value provided by each channel. In this regard, MMM has matured to become a staple approach for cross-channel media measurement. Leading tech entities like Google and Meta have significantly contributed to advancing Marketing Mix Modeling (MMM), introducing their respective open-source packages: Lightweight [1] and Robyn [2]. Lightweight employs Bayesian statistics to incorporate business domain knowledge into media priories [3], while Robyn utilizes the frequentist approach and multiple objective functions, generating thousands of iterations. But it is often an arduous task to further disentangle partner-level contributions that can provide actionable, in-depth insights that are required for media planning and budgeting decisions. Consider an instance where the MMM highlights television (TV) as a high-performance channel. The pressing query that follows is: can we disentangle the TV partner-level (e.g., Disney, ESPN) performance to potentially reallocate investments from the underperformers to the stronger ones?

We propose a structured Designated Market Area (DMA) test as a potential approach. Inspired by cooperative game theory and first published by Shapley [4], the test can be designed in a way that makes it possible to obtain the precise partner-level contributions. But the DMA test can prove complicated if not impossible to implement especially when the number of partners is large. Alternatively, Shapley Value Regression presents itself as a viable approach to measure partner-level performance without the need of experimental tests and consequently, effectively transforming the attribution problem into a variable importance problem within regression models. A case with real data from the financial service industry is given to illustrate how Shapley Value Regression is applied. Furthermore, a straightforward method for deriving co-efficient is proposed and compared against other proposed approaches. The python program in Jupyter Notebook is attached in the appendix.

# 2. Structured DMA Test with Shapley Value

## 2.1 Shapley Value

The theory behind Shapley Values as presented by Shapley [4] is based on cooperative game theory where a set of players, referred to as a coalition, are playing towards a common goal. The value added by each player towards achieving the goal can be quantified by considering all possible combinations of players against the outcome and thereby assessing their marginal contribution. Shapley Values are calculated as the weighted average of these marginal contributions and since the players are put through all possible scenarios, the credit attributed by averaging these values is mathematically and intuitively fair.

Shapley Value is calculated as follows:

$$\varphi_i(v) = \sum_{S \subseteq N \setminus \{i\}} \frac{|S|!\,(n - |S| - 1)!}{n!} \big(v(S \cup \{i\}) - v(S)\big) \qquad (1)$$

$\varphi_i$: Shapley Value of channel i

N: Total number of partners within channel i

S: Coalitions excluding channel i

v: utility function that represents contributions from a coalition

## 2.2 DMA Tests Implementation with Shapley Value

In a hypothetical scenario that MMM indicates the TV channel is driving $10MM sales/year with commercials aired on Disney, ESPN and CNN. The goal is to disentangle and measure the impact of each TV partner. A structured DMA test inspired by game-theory is proposed to solve the problem.

In this case the players are TV partners (Disney, ESPN and CNN). But applying the game theory approach involves more than just implementing the Shapley Value equation. Often, the definition and collection of input data may play a bigger role. To accurately calculate the contributions of each coalition, the most ideal way is through tests. Nielsen coined the term DMA [5] that divides US market into 210 areas and has become a popular way to split test and control geo-graphically in the cross-channel environment. In this case, DMA is the only feasible way to split test and control groups. Here are the steps:

Step 1. Identify 8 comparable DMAs with similar population profiles.

Step 2. Conduct the DMA test and track the sales by DMA

| DMA | Partners Aired | Sales |
|---|---|---|
| 1 | NA (baseline sales) | $ 1000 |
| 2 | Disney | $ 10,000 |
| 3 | ESPN | $ 5,000 |
| 4 | CNN | $ 12,000 |
| 5 | ESPN, CNN | $ 15,000 |
| 6 | Disney, ESPN | $ 12,000 |
| 7 | Disney, CNN | $ 20,000 |
| 8 | Disney, ESPN, CNN | $ 25,000 |

Step 3. Calculate the Shapley Value for Disney based on the Shapley Value equation (1):

| Coalitions | Sales | Coalitions without Disney | Sales without Disney | Marginal Contributions from Disney | Shapley Value Weight | Shapley Value |
|---|---|---|---|---|---|---|
| Disney | $ 10,000 | NA | $ 1,000 | $ 9,000 | 0.33 | $ 3,000 |
| Disney, ESPN | $ 12,000 | ESPN | $ 5,000 | $ 7,000 | 0.17 | $ 1,167 |
| Disney, CNN | $ 20,000 | CNN | $ 12,000 | $ 8,000 | 0.17 | $ 1,333 |
| Disney, ESPN, CNN | $ 25,000 | ESPN,CNN | $ 15,000 | $ 10,000 | 0.33 | $ 3,333 |
| | | | | | **Shapley Value for Disney** | **$ 8,833** |

Step 4. Repeat the same process for ESPN and CNN

| Coalitions | Sales | Coalitions without ESPN | Sales without ESPN | Marginal Contributions from ESPN | Shapley Value Weight | Shapley Value |
|---|---|---|---|---|---|---|
| ESPN | $ 5,000 | NA | $ 1,000 | $ 4,000 | 0.33 | $ 1,333 |
| ESPN, Disney | $ 12,000 | Disney | $ 10,000 | $ 2,000 | 0.17 | $ 333 |
| ESPN, CNN | $ 15,000 | CNN | $ 12,000 | $ 3,000 | 0.17 | $ 500 |
| Disney, ESPN, CNN | $ 25,000 | Disney,CNN | $ 20,000 | $ 5,000 | 0.33 | $ 1,667 |
| | | | | | **Shapley Value for ESPN** | **$ 3,833** |

| Coalitions | Sales | Coalitions without CNN | Sales without CNN | Marginal Contributions from CNN | Shapley Value Weight | Shapley Value |
|---|---|---|---|---|---|---|
| CNN | $ 12,000 | NA | $ 1,000 | $ 11,000 | 0.33 | $ 3,667 |
| CNN, ESPN | $ 15,000 | ESPN | $ 5,000 | $ 10,000 | 0.17 | $ 1,667 |
| CNN, Disney | $ 20,000 | Disney | $ 10,000 | $ 10,000 | 0.17 | $ 1,667 |
| Disney, ESPN, CNN | $ 25,000 | Disney, ESPN | $ 12,000 | $ 13,000 | 0.33 | $ 4,333 |
| | | | | | **Shapley Value for CNN** | **$ 11,333** |

Step 5. Calculate the % contribution

| Partners | Shapley Value | %Contribution |
|---|---|---|
| ESPN | $ 3,833 | 16% |
| Disney | $ 8,833 | 37% |
| CNN | $ 11,333 | 47% |
| Total | $ 24,000 | 100% |

Step 6.

Extrapolate the DMA test result to the national level. There we can derive the $ contribution by partner.

| Partners | %Contribution | $Contribution |
|---|---|---|
| ESPN | 16% | $ 1,597,222 |
| Disney | 37% | $ 3,680,556 |
| CNN | 47% | $ 4,722,222 |
| Total | 100% | $ 10,000,000 |

Although testing is considered the golden standard to detect the casual relationship due to concerns of the Simpson's paradox [6], there are unique challenges in this space:

1. The number of testing DMAs will increase exponentially as the number of partners increases. For example, 6 partners require 64 comparable DMAs. The execution of such test is complicated and prone to error, not to mention 64 comparable DMAs are impossible to find.
2. Unlike in-channel creative testing, certain cross-contaminations between testing and control DMAs are inevitable as people travel across DMAs.

Hence, the Shapley Value Regression is a more practical approach to address the problem.

# 3. Shapley Value Regression

Expanding upon the game-theory and in the context of regression, Shapley Value Regression, initially proposed by Lindeman et al. [7] and later popularly presented by Kruskal [8][9], treats the independent variables in the regression analysis as players that collaboratively aim to predict the dependent variable, mirroring the cooperative nature of a game.

Shapley Value Regression stands out in the cross-channel partner measurement domain due to several key characteristics:

1. The sum of total partner contributions will always sum up to the channel contribution.
2. A partner that makes no marginal contribution to any coalition will get zero credit but not negative credit. This is aligned with the prevalent assumption that unlike political marketing, commercial marketing efforts typically yield a positive impact on sales or, at the least, maintain status quo, rather than negatively affecting outcomes.
3. It mitigates multicollinearity problem, which is commonly seen in cross-channel-partner marketing.

The following sub-sections detail the Shapley Value Regression approach to better understand partner efficiencies through a real-world data set sourced from the financial services industry.

## 3.1 Understanding the Data

Similar to the hypothetical example, TV is the marketing channel under study in the following example and partners are the different TV network partners where advertisements were aired from the period of July 2020 to December 2022. Gross rating point (GRP) is used to represent the viewership of each partner and is considered the independent variables. The dependent variable is the attributed TV sales from MMM. To re-iterate the problem, the goal is to further attribute the overall TV sales to each TV partner.

*Table 1. A snapshot of the time-series data with the dependent and independent variables*

| Week | Partner A | Partner B | Partner C | Partner D | Partner E | TV Sales from MMM |
|---|---|---|---|---|---|---|
| 6/1/2020 | 1.713 | -0.677 | 0.100 | -0.458 | 1.109 | 0.520 |
| 6/8/2020 | 1.995 | -0.973 | 0.210 | -0.471 | 1.172 | 0.607 |
| 6/15/2020 | 2.145 | -0.811 | 0.147 | -0.398 | 0.977 | 0.667 |
| 6/22/2020 | 2.276 | -0.940 | -0.031 | -0.292 | 0.839 | 0.636 |
| 6/29/2020 | -0.973 | -0.973 | -0.973 | -0.973 | -0.973 | -0.773 |
| 7/6/2020 | -0.973 | -0.973 | -0.836 | -0.973 | -0.973 | -1.176 |
| 7/13/2020 | 1.168 | -0.818 | 0.272 | 0.153 | 1.704 | 0.073 |
| 7/20/2020 | 1.216 | 0.260 | 0.498 | 1.489 | 1.115 | 1.072 |
| 7/27/2020 | 1.882 | -0.486 | 0.133 | 0.871 | 1.244 | 1.168 |
| 8/3/2020 | 3.301 | -0.756 | -0.001 | 0.494 | 1.825 | 1.582 |
| 8/10/2020 | 4.677 | -0.268 | 0.433 | 0.576 | 1.578 | 2.320 |
| 8/17/2020 | 1.693 | -0.407 | 0.289 | 1.078 | 1.360 | 1.724 |
| 8/24/2020 | -0.973 | -0.787 | -0.970 | -0.973 | -0.973 | -0.412 |
| 8/31/2020 | 1.027 | -0.604 | 1.502 | 0.662 | 0.645 | 0.653 |
| 9/7/2020 | 1.241 | -0.569 | 1.727 | 2.068 | 0.710 | 1.503 |
| 9/14/2020 | 0.699 | -0.544 | 1.749 | 2.437 | 0.544 | 1.696 |
| 9/21/2020 | -0.973 | -0.973 | -0.973 | -0.973 | -0.973 | -0.500 |
| 9/28/2020 | -0.851 | -0.973 | -0.645 | -0.973 | -0.973 | -0.969 |
| 10/5/2020 | 0.147 | -0.593 | -0.204 | 1.071 | 0.061 | -0.282 |
| 10/12/2020 | 0.540 | 0.011 | -0.849 | 1.178 | 0.234 | 0.079 |
| 10/19/2020 | 0.715 | 0.602 | -0.664 | 0.817 | 0.417 | 0.398 |

*Note: Data has been scaled and normalized to keep confidentiality*

It is worthy to note that if the MMM was built at a channel-partner level instead of the channel level, performing Shapley Value Regression could have been avoided. But since MMM will be at high risk of overfitting if all the media channels are at partner level, Shapley Value Regression can prove to be extremely beneficial in such scenarios.

### 3.2 Relative Importance of Marketing Partners

$R^2$, or the coefficient of determination, is the proportion of the variation in the dependent variable that is explained by the independent variables. As proposed by Lindeman et al. [7], the relative importance of marketing partners is calculated by performing linear regression on every possible coalitions of independent variables and averaging the marginal increase in $R^2$ values.

Table 2 details all the possible coalitions of the predictors and the corresponding $R^2$ resulting from their linear regressions. For the use case being discussed in this paper, the intercept in linear regression is forced to zero with the business intuition that the TV sales indicated by MMM would be 100% attributed back to TV partners within that channel.

*Table 2. All coalitions of Partner A, B, C, D and E*

| S. No. | Coalition | $R^2$ |
|---|---|---|
| 0 | [] | 0 |
| 1 | [Partner A] | 0.562 |
| 2 | [Partner A, Partner B] | 0.647 |
| 3 | [Partner A, Partner C] | 0.751 |
| 4 | [Partner A, Partner D] | 0.756 |
| 5 | [Partner A, Partner E] | 0.739 |
| 6 | [Partner A, Partner B, Partner C] | 0.799 |
| 7 | [Partner A, Partner B, Partner D] | 0.793 |
| 8 | [Partner A, Partner B, Partner E] | 0.812 |
| 9 | [Partner A, Partner C, Partner D] | 0.843 |
| 10 | [Partner A, Partner C, Partner E] | 0.818 |
| 11 | [Partner A, Partner D, Partner E] | 0.793 |
| 12 | [Partner A, Partner B, Partner C, Partner D] | 0.870 |
| 13 | [Partner A, Partner B, Partner C, Partner E] | 0.869 |
| 14 | [Partner A, Partner B, Partner D, Partner E] | 0.840 |
| 15 | [Partner A, Partner C, Partner D, Partner E] | 0.856 |
| 16 | [Partner A, Partner B, Partner C, Partner D, Partner E] | 0.889 |

The following step calculates the Shapley Value of each partner using the formula below.

Let there be $n$ channel partners $\{x_1, x_2, \ldots, x_n\}$ working in a coalition $N$, the Shapley Value of an individual partner $i$, represented by $\varphi_i(v)$, can be calculated as:

$$\varphi_i(v) = \sum_{S \subseteq N \setminus \{i\}} \frac{|S|!\,(n - |S| - 1)!}{n!} \left(v(S \cup \{i\}) - v(S)\right) \quad (1)$$

In this formula, marginal contribution is the value added by each partner which is calculated as:
$$(v(S \cup \{i\}) - v(S))$$

Since $R^2$ is the chosen utility function in this analysis, the marginal contribution of a partner(s) can be calculated as the increments in $R^2$ value of the coalition that contains the partner compared to the coalition where that partner is absent.

*Table 3. The marginal contribution for Partner A.*

| S. No. | Coalition | Coalition Excluding Partner A | $R^2$ of Coalition | $R^2$ of Coalition Excluding Partner A | Marginal Contribution from Partner A |
|---|---|---|---|---|---|
| 1 | [Partner A] | [] | 0.562 | 0.000 | 0.562 |
| 2 | [Partner A, Partner B] | [Partner B] | 0.647 | 0.281 | 0.366 |

| 3 | [Partner A, Partner C] | [Partner C] | 0.751 | 0.559 | 0.192 |
| 4 | [Partner A, Partner D] | [Partner D] | 0.756 | 0.510 | 0.247 |
| 5 | [Partner A, Partner E] | [Partner E] | 0.739 | 0.605 | 0.134 |
| 6 | [Partner A, Partner B, Partner C] | [Partner B, Partner C] | 0.799 | 0.656 | 0.142 |
| 7 | [Partner A, Partner B, Partner D] | [Partner B, Partner D] | 0.793 | 0.600 | 0.193 |
| 8 | [Partner A, Partner B, Partner E] | [Partner B, Partner E] | 0.812 | 0.731 | 0.081 |
| 9 | [Partner A, Partner C, Partner D] | [Partner C, Partner D] | 0.843 | 0.716 | 0.127 |
| 10 | [Partner A, Partner C, Partner E] | [Partner C, Partner E] | 0.818 | 0.732 | 0.086 |
| 11 | [Partner A, Partner D, Partner E] | [Partner D, Partner E] | 0.793 | 0.662 | 0.130 |
| 12 | [Partner A, Partner B, Partner C, Partner D] | [Partner B, Partner C, Partner D] | 0.870 | 0.767 | 0.102 |
| 13 | [Partner A, Partner B, Partner C, Partner E] | [Partner B, Partner C, Partner E] | 0.869 | 0.812 | 0.056 |
| 14 | [Partner A, Partner B, Partner D, Partner E] | [Partner B, Partner D, Partner E] | 0.840 | 0.754 | 0.085 |
| 15 | [Partner A, Partner C, Partner D, Partner E] | [Partner C, Partner D, Partner E] | 0.856 | 0.769 | 0.087 |
| 16 | [Partner A, Partner B, Partner C, Partner D, Partner E] | [Partner B, Partner C, Partner D, Partner E] | 0.889 | 0.829 | 0.060 |

$$\frac{|S|!\,(n - |S| - 1)!}{n!}$$

This part of the equation can be understood as the weight. It is calculated as follows:

Table 4. The calculated weights for Partner A.

| S. No. | Coalition | Coalition Excluding Partner A | S | N | $\frac{|S|!\,(n - |S| - 1)!}{n!}$ |
| --- | --- | --- | --- | --- | --- |
| 1 | [Partner A] | [] | 0 | 5 | 0.2 |
| 2 | [Partner A, Partner B] | [Partner B] | 1 | 5 | 0.05 |
| 3 | [Partner A, Partner C] | [Partner C] | 1 | 5 | 0.05 |
| 4 | [Partner A, Partner D] | [Partner D] | 1 | 5 | 0.05 |
| 5 | [Partner A, Partner E] | [Partner E] | 1 | 5 | 0.05 |
| 6 | [Partner A, Partner B, Partner C] | [Partner B, Partner C] | 2 | 5 | 0.033333 |
| 7 | [Partner A, Partner B, Partner D] | [Partner B, Partner D] | 2 | 5 | 0.033333 |
| 8 | [Partner A, Partner B, Partner E] | [Partner B, Partner E] | 2 | 5 | 0.033333 |
| 9 | [Partner A, Partner C, Partner D] | [Partner C, Partner D] | 2 | 5 | 0.033333 |
| 10 | [Partner A, Partner C, Partner E] | [Partner C, Partner E] | 2 | 5 | 0.033333 |
| 11 | [Partner A, Partner D, Partner E] | [Partner D, Partner E] | 2 | 5 | 0.033333 |
| 12 | [Partner A, Partner B, Partner C, Partner D] | [Partner B, Partner C, Partner D] | 3 | 5 | 0.05 |
| 13 | [Partner A, Partner B, Partner C, Partner E] | [Partner B, Partner C, Partner E] | 3 | 5 | 0.05 |
| 14 | [Partner A, Partner B, Partner D, Partner E] | [Partner B, Partner D, Partner E] | 3 | 5 | 0.05 |

| 15 | [Partner A, Partner C, Partner D, Partner E] | [Partner C, Partner D, Partner E] | 3 | 5 | 0.05 |
| 16 | [Partner A, Partner B, Partner C, Partner D, Partner E] | [Partner B, Partner C, Partner D, Partner E] | 4 | 5 | 0.2 |

From equation (1) the Shapley Value can now be calculated as the product of weight and marginal contribution summed over all the coalitions that Partner A belongs to.

Table 5 shows the calculation of the Shapley Value for Partner A. The Shapley Values for the rest of the partners are calculated in a similar fashion and the results are shown in Table 6.

*Table 5. Shapley Values for Partner A*

| S. No. | Coalition | Coalition Excluding Partner A | $\frac{\lvert S\rvert!\,(n - \lvert S\rvert - 1)!}{n!}$ | Marginal Contribution | Shapley Value |
|---|---|---|---|---|---|
| 1 | [Partner A] | [] | 0.2 | 0.562 | **0.1123** |
| 2 | [Partner A, Partner B] | [Partner B] | 0.05 | 0.366 | **0.0183** |
| 3 | [Partner A, Partner C] | [Partner C] | 0.05 | 0.192 | **0.0096** |
| 4 | [Partner A, Partner D] | [Partner D] | 0.05 | 0.247 | **0.0123** |
| 5 | [Partner A, Partner E] | [Partner E] | 0.05 | 0.134 | **0.0067** |
| 6 | [Partner A, Partner B, Partner C] | [Partner B, Partner C] | 0.033333 | 0.142 | **0.0047** |
| 7 | [Partner A, Partner B, Partner D] | [Partner B, Partner D] | 0.033333 | 0.193 | **0.0064** |
| 8 | [Partner A, Partner B, Partner E] | [Partner B, Partner E] | 0.033333 | 0.081 | **0.0027** |
| 9 | [Partner A, Partner C, Partner D] | [Partner C, Partner D] | 0.033333 | 0.127 | **0.0042** |
| 10 | [Partner A, Partner C, Partner E] | [Partner C, Partner E] | 0.033333 | 0.086 | **0.0029** |
| 11 | [Partner A, Partner D, Partner E] | [Partner D, Partner E] | 0.033333 | 0.130 | **0.0043** |
| 12 | [Partner A, Partner B, Partner C, Partner D] | [Partner B, Partner C, Partner D] | 0.05 | 0.102 | **0.0051** |
| 13 | [Partner A, Partner B, Partner C, Partner E] | [Partner B, Partner C, Partner E] | 0.05 | 0.056 | **0.0028** |
| 14 | [Partner A, Partner B, Partner D, Partner E] | [Partner B, Partner D, Partner E] | 0.05 | 0.085 | **0.0043** |
| 15 | [Partner A, Partner C, Partner D, Partner E] | [Partner C, Partner D, Partner E] | 0.05 | 0.087 | **0.0043** |
| 16 | [Partner A, Partner B, Partner C, Partner D, Partner E] | [Partner B, Partner C, Partner D, Partner E] | 0.2 | 0.060 | **0.0121** |

| Shapley Values for Partner A | 0.213 |
|---|---|

The Shapley Values shown in Table 6 provide an idea of the relative importance of each of the independent variables in predicting the target variable. The Shapley Values can be normalized to find the share of sales driven by each partner.

*Table 6. Shapley Value for all partners*

| S. No. | Parent Owner | Shapley Value | Normalized Share |
|---|---|---|---|
| 1 | Partner A | 0.213 | 24% |
| 2 | Partner B | 0.105 | 12% |
| 3 | Partner C | 0.203 | 23% |
| 4 | Partner D | 0.166 | 19% |
| 5 | Partner E | 0.203 | 23% |
|   | **Total** | **0.889** | **100%** |

From Table 6, it is also observed that Shapley Values of each partner obtained by using the Shapley Value Regression method are always positive if not 0, and total to the $R^2$ when all the partners work in a coalition.

This inherent property of Shapley Values is very useful when deriving partner-level insights from a highly correlated dataset by providing the ability to isolate the performance of each partner.

The normalized Shapley share $\varphi_i(v)_{norm}$ for the $i^{th}$ channel partner obtained from above can be multiplied with the response variable column ($Y$) to determine the sales attributed to that partner, represented by $O$.

$$O_i = \sum(Y) * \varphi_i(v)_{norm} \qquad (2)$$

Leveraging any additional information such as the marketing spend on that partner and the GRPs received by them, it is possible to arrive at an estimated cost per sales and the partner efficiencies can be determined in the method that is beneficial for the use case.

## 4. Calculating Coefficients in Shapley Value Regression

The coefficients are not necessary to quantify the overall partner level attribution. But if the coefficients are available, it is beneficial to

1. plot the week-over-week predicted values and visually see how well they fit against the actual data.

2. forecast future channel-partner level sales

The best approach to calculate the co-efficient is being debated. Lipovetsky [11] found the Sharply Value net effect and adjusted the coefficients by decomposing the $R^2$ and forming a quadratic equation to minimize its square distance with Sharply Value. Building on that approach, Mishra [10] provided different optimization techniques such as Host-Parasite Co-Evolutionary (HPC).

Gromping and Landau [12] criticized Lipovetsky's approach being misleading in certain cases. Their central argument is if there is a strong negative causal relationship between an independent variable and dependent variable, Lipovetsky's approach can assign a large positive coefficient. Gromping then offered an OLS with non-negativity constraints to possibly trim the variable when the relationship is negative. However, we believe that Gromping's argument rather reveals the fundamental flaw of Shapley Value Regression itself. His non-negativity constraints approach is nothing related to the game theory and will suffer the same multi-collinearity issue as the regular regression.

In this paper we propose a very straightforward approach to calculate the coefficients in Shapley Value Regression. The regression equation can be written as

$$Y = \sum_1^n (\beta_i * \sum_1^m x_{i,j}) \tag{3}$$

$\beta_i$ is the coefficient for player $x_i$

$i$ is the ith player

$n$ is the total number of players

$j$ is the jth data points in the time-series

$m$ is the total number of data points in the time-series

Note there is no intercept in the equation assuming Y is contributed 100% by the sum of $x_i$. Given that total contributed sales by partner i has been calculated as $O_i$, the regression equation can also be written as

$$Y = \sum_1^n O_i \tag{4}$$

Given (3) and (4), the beta coefficients could be technically calculated as:

$$\beta_i = \frac{O_i}{\sum_1^m x_{i,j}}$$

## 4.1 Comparison of Co-efficient Deduction Approaches in Shapley Value Regression

In Mishra's paper [10], the author calculated the coefficients using Shapley Values by using quadratic optimization.

Mishra used a dummy dataset and compared linear regression and the coefficients produced by HPC. The dataset is below:

| Table 7. The Portland Cement Dataset (cf. Woods, Steinour and Starke, 1932) | | | | | | | | | | | | | |
|---|---|---|---|---|---|---|---|---|---|---|---|---|---|
| | 1 | 2 | 3 | 4 | 5 | 6 | 7 | 8 | 9 | 10 | 11 | 12 | 13 |
| $x_1$ | 7 | 1 | 11 | 11 | 7 | 11 | 3 | 1 | 2 | 21 | 1 | 11 | 10 |
| $x_2$ | 26 | 29 | 56 | 31 | 52 | 55 | 71 | 31 | 54 | 47 | 40 | 66 | 68 |
| $x_3$ | 6 | 15 | 8 | 8 | 6 | 9 | 17 | 22 | 18 | 4 | 23 | 9 | 8 |
| $x_4$ | 60 | 52 | 20 | 47 | 33 | 22 | 6 | 44 | 22 | 26 | 34 | 12 | 12 |
| y | 78.5 | 74.3 | 104.3 | 87.6 | 95.9 | 109.2 | 102.7 | 72.5 | 93.1 | 115.9 | 83.8 | 113.3 | 109.4 |

Using the dataset above, we found the coefficients using our approach and compared them to Mishra's paper and linear regression (OLS).

| Table 8. Regression Coefficients by Method | | | | | |
|---|---|---|---|---|---|
| | Co-efficients | | | | Intercept/Constant |
| | $x_1$ | $x_2$ | $x_3$ | $x_4$ | |
| Our Approach | 3.24006 | 0.5875 | 1.11326 | 0.99515 | 0 |
| HPC - Regular | 0.82883 | 0.33204 | -0.6289 | -0.3136 | 90.1 |
| OLS | 1.551 | 0.51 | 0.102 | -0.144 | 62.4 |

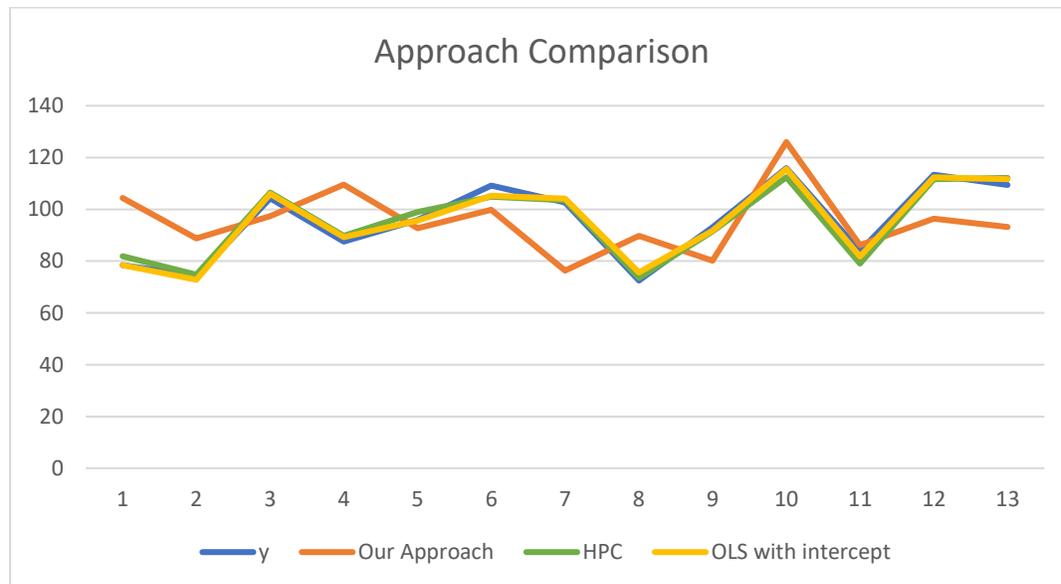

From there, we compared how each coefficient performed when predicting values. The blue line is the actual dependent variable. The HPC approach is close to OLS, both of which unarguably fit the actual dependent variable better. Although our approach underperforms HPC and OLS for fitness, both HPC and OLS would suffer interpretability as the negative co-efficient ($x_3, x_4$) violates the basic assumption in our use case, which is marketing partners will not have a negative impact on business KPIs. The HPC approach surprisingly gave negative co-efficient despite it is based on

Shapley Value Regression, because the cost function proposed by Lipovetsky is trying to minimize the sum of square difference of the Shapley Value of each player, and the contribution of each player to $R^2$, but it doesn't guarantee the Shapley Value of each player will remain positive. The approach that Gromping proposed will not only suffer multi-collinearity, but is also not practical in the business context when multiple partners (especially the top invested partners) are trimmed due to the negative sign, indicating none of them contributing anything to the channel.

It is possible to improve Lipovetsky's approach with additional restrictions to make it more practical in the marketing measurement space:

1. The sign of each coefficient >0
2. The magnitude of each coefficient should be in the same order as the Shapley Value

However, with these restrictions that aim to improve the interpretability, the $R^2$ or the goodness of fit is likely to be compromised.

## 5. A Python Program in Jupyter Notebook

A Python program in Jupyter Notebook is attached in the Appendix. The input data is a time-series data that is stored as an Excel file with column headers. The first column is the date (in weeks), the last column is the dependent variable while the rest are independent variables. Only basic python packages are used to produce the Shapley Value Regression and the co-efficient following the steps illustrated in the above section. Please note this is not to be confused with packages like SHAP that aim to explain the variable importance in machine learning models.

## 6. Conclusion

In conclusion, Shapley Value Regression offers a promising avenue to unravel partner-level performance within marketing channels. Its ability to dissect contributions without succumbing to multicollinearity presents a valuable tool for marketers. The discussion on coefficient calculation methods underscores the need for a balanced approach between interpretability and model accuracy.

## Appendix

```python
#package installations

import pandas as pd

import numpy as np

import itertools

import math

from matplotlib import pyplot as plt

import sklearn

from sklearn.linear_model import LinearRegression

from sklearn.preprocessing import StandardScaler

#function to read input data

def read_data(path, sheet_name):

    input_data = pd.read_excel(path, sheet_name=sheet_name, engine='openpyxl')

    input_data = input_data.loc[:, ~input_data.columns.str.contains('^Unnamed')]

    input_data.fillna(0, inplace = True)

    return input_data

#function to form all combinations of independent variables

def form_combinations(all_columns, dependent_var, excluded_vars):

    independent_vars = list(set(all_columns) - set(excluded_vars) - set(dependent_var))

    independent_vars.sort()

    all_combinations = []

    for var in range(len(independent_vars) + 1):

        for var_subset in itertools.combinations(independent_vars, var):

            all_combinations.append(var_subset)

    return independent_vars, all_combinations
```

```python
#function to run Linear Regression for each combination of independent variables
def form_coalitions(X, y, x_var_combinations):
    coalitions = pd.DataFrame(columns=["Independent Variables", "R Squared", "Coefficients"])

    for independent_var_combo in x_var_combinations:
        if independent_var_combo != ():
            X_curr = X[list(independent_var_combo)]
            scaler = StandardScaler()
            scaler.fit(X_curr)
            X_std = scaler.transform(X_curr)
            X_std = pd.DataFrame(X_std)
            X_std.columns = X_curr.columns

            scaler.fit(pd.DataFrame(y))
            Y_std = scaler.transform(pd.DataFrame(y))

            model = LinearRegression(fit_intercept=False).fit(X_std,Y_std)
            r_sq = model.score(X_std, Y_std)
            coeff = model.coef_

            coalition_result = {
            "Independent Variables": list(independent_var_combo),
            "R Squared": r_sq, "Coefficients": coeff
            }
            coalitions = coalitions.append(coalition_result,
                                            ignore_index = True)

        else:
            r_sq = 0
```

```python
                coalition_result = {
                "Independent Variables": list(independent_var_combo),
                "R Squared": r_sq, "Coefficients": [0]
                }
                coalitions = coalitions.append(coalition_result,
                                                ignore_index = True)

    coalitions['R Squared'] = coalitions['R Squared'].astype(float)

    return coalitions, X_std, Y_std

#function to calculate Shapley value for each coalition
def get_shapley_values(independent_vars, coalitions):
    shapley_values = pd.DataFrame(columns = ["Coalition", "Current Partner",
                        "Other Partners", "R Squared", "S", "N",
            "Weight", "Marginal Contribution", "Shapley Value"])
    for var in independent_vars:
        for index, row in coalitions.iterrows():

                var_combo_list = row['Independent Variables']
                if var in row['Independent Variables']:
                        other_channels = list(filter(lambda x: x not in
                                        set([var]),var_combo_list))
                        r_sq = row['R Squared']
                        s = len(var_combo_list) - 1
                        n = len(independent_vars)
                        weight = math.factorial(abs(s))*math.factorial(n-
                                        abs(s)-1)/math.factorial(n)
                        marginal_contribution = 0
                        shapley_value = 0
                        shapley_values = shapley_values.append({'Coalition':
                                var_combo_list, 'Current Partner': var,
```

```python
                                        'Other Partners': other_channels,'R Squared':
                                        r_sq, 'S': s, 'N':n , "Weight": weight, 'Marginal
                                        Contribution': marginal_contribution, 'Shapley
                                        Value': shapley_value}, ignore_index = True)

    for shapley_index, shapley_row in shapley_values.iterrows():
        for coalition_index, coalition_row in coalitions.iterrows():
            if shapley_row['Other Partners'] == coalition_row['Independent
                                                Variables']:
                marginal_contribution = shapley_row['R Squared'] -
                                        coalition_row['R Squared']
                shapley_values.loc[shapley_index, 'Marginal Contribution']
                                        = marginal_contribution
                shapley_values.loc[shapley_index, 'Shapley Value'] =
                                shapley_row['Weight'] * marginal_contribution
    return shapley_values

#function to get Shapley adjusted and regular Linear Regression coefficients
def get_coefficients(shapley_values, y, X, dependent_var):
    total_shap = shapley_values['Shapley Value'].sum()
    norm_shapley_values = shapley_values.groupby("Current Partner")[['Shapley
                                                    Value']].sum()
    norm_shapley_values['Normalized Shapley Value'] = norm_shapley_values['Shapley
                                            Value']/total_shap*100
    norm_shapley_values = norm_shapley_values.reset_index()
    norm_shapley_values = norm_shapley_values.rename(columns={'Current Partner':
                                                'Parent Owner'})
    total_y = sum(y[dependent_var])
    norm_shapley_values["Reattributed Visits"] = norm_shapley_values['Normalized
                                        Shapley Value']*total_y/100
    reattr_values = norm_shapley_values['Normalized Shapley Value']*total_y/100

    shap_reg_coeff = []
    k=0
    for i in X_train.columns:
```

```python
        temp = []
        for j in range(0, len(reattr_values)):
            i_sum = sum(X[i])
            temp.append(reattr_values[j]/i_sum)
        shap_reg_coeff.append(temp[k])
        k=k+1
    norm_shapley_values["Shapley Coefficients"] = shap_reg_coeff
    
    lin_reg = LinearRegression(fit_intercept = False).fit(X, y)
    lin_reg_r = lin_reg.score(X,y)
    lin_reg_coeff = lin_reg.coef_
    
    norm_shapley_values["Regression Coefficients"] = lin_reg_coeff[0]
    
    return norm_shapley_values
```